\documentclass[fleqn,10pt]{wlscirep}
\usepackage[utf8]{inputenc}
\usepackage[T1]{fontenc}
\usepackage{listings}
\usepackage{tabularx}
\usepackage{booktabs}
\usepackage{listings}
\usepackage{xcolor}

\title{
Deep Fast Vision: A Python Library for Accelerated Deep Transfer Learning Vision Prototyping}

\author[1,2,*]{Fabi Prezja}
\affil[1]{University of Jyväskylä, Faculty of Information Technology, Jyväskylä, Finland}
\affil[2]{Finnish Artificial Intelligence Research Network, Jyväskylä, Finland}
\affil[*]{corresponding.faprezja@jyu.fi}

\begin{abstract}
Deep learning-based vision is characterized by intricate frameworks that often necessitate a profound understanding, presenting a barrier to newcomers and limiting broad adoption. With many researchers grappling with the constraints of smaller datasets, there's a pronounced reliance on pre-trained neural networks, especially for tasks such as image classification. This reliance is further intensified in niche imaging areas where obtaining vast datasets is challenging. Despite the widespread use of transfer learning as a remedy to the small dataset dilemma, a conspicuous absence of tailored auto-ML solutions persists. Addressing these challenges is "Deep Fast Vision", a python library that streamlines the deep learning process. This tool offers a user-friendly experience, enabling results through a simple nested dictionary definition, helping to democratize deep learning for non-experts. Designed for simplicity and scalability, Deep Fast Vision appears as a bridge, connecting the complexities of existing deep learning frameworks with the needs of a diverse user base.
\end{abstract}
\begin{document}

\flushbottom
\maketitle
%
%
\thispagestyle{empty}

\section*{Introduction}

In the rapidly evolving realm of artificial intelligence, deep learning\cite{lecun2015deep} has cemented its position as a transformative technology, fundamentally reshaping various sectors, from medicine and well-being \cite{kather2019predicting,bychkov2018deep,skrede2020deep,calimeri2017biomedical,prezja2022deepfake,prezja2022h,prezja2023improved,prezja2023improving} to finance\cite{huang2020deep}. Central to this revolution is the domain of vision-based deep learning, which has enabled machines to perceive, interpret, and act upon visual data with unparalleled accuracy. However, as with any new technology, the path to its widespread adoption is riddled with challenges. One of the predominant hurdles lies in the intricate nature of current deep learning frameworks. These platforms, while powerful, often require an in-depth understanding and expertise, thus posing a daunting barrier to entry for newcomers or those without specialized training. This complexity not only hinders accessibility but also stifles broad-based adoption, preventing a larger community from harnessing the potential of deep learning in vision tasks.

Further complicating the landscape is the inherent challenge of acquiring vast, labeled datasets, especially in niche imaging areas. Many researchers and developers find themselves constrained by the limited data at their disposal. As a result, there's a pronounced shift towards leveraging pre-trained neural networks, which, while beneficial, are not without thier own set of challenges and limitations.

Amidst these challenges, the technique of transfer learning \cite{zhuang2020comprehensive} has emerged, offering a pathway to leverage knowledge from extensive datasets to enhance performance on smaller, more specific datasets. Yet, even this approach is not without its hurdles, notably the glaring absence of comprehensive auto-ML solutions tailored for vision-based transfer learning. This paper introduces "Deep Fast Vision \cite{fabprezjad_2023}," a Python library designed to address these multifaceted challenges. By streamlining processes and reducing the steep learning curve associated with contemporary frameworks, it aims to democratize deep learning in vision, making it accessible and efficient for both novices and experts. In the subsequent sections, we will delve deeper into the challenges, explore the nuances of "Deep Fast Vision," and elucidate how it stands as a tool in bridging the gap between complex frameworks and broad user-centric solutions.

\section*{Methods}

\subsection*{Automated Machine Learning Repository}

Automated Machine Learning (Auto-ML) simplifies the comprehensive process of implementing machine learning to practical applications. In scenarios where specialized knowledge is limited yet, there's an urgent need to extract insights from data, Auto-ML proves invaluable. Regarding transfer learning auto-ml in vision, two prominent libraries emerge AutoKeras\cite{JMLR:v24:20-1355} and Deep Fast Vision\cite{fabprezja_2023}.

\subsection*{AutoKeras}

AutoKeras\cite{JMLR:v24:20-1355} stands out as a freely accessible Auto-ML solution for deep learning, constructed on the foundation of Keras. Its prowess lies in its ability to autonomously pinpoint the most fitting deep learning model suited to a dataset. Integrated features of this library encompass automatic model identification and hyperparameter optimization. It's versatile in handling diverse data formats, encompassing images, textual content, and organized data.

\subsection*{Deep Fast Vision}

Deep Fast Vision\cite{fabprezjad_2023} is a Python library designed to facilitate the rapid prototyping of deep transfer learning models for vision tasks. It's built on the foundation of TensorFlow and Keras, two leading deep learning frameworks. The primary objective of Deep Fast Vision is to make the process of model prototyping more accessible and faster for both beginners and experts in the field.

\subsubsection*{Configuration}

The library allows users to run their computations either on a CPU or GPU, although the use of a GPU is recommended to leverage the acceleration benefits it provides for deep learning tasks.

\subsubsection*{Parameter Overview}

To streamline the prototyping process, Deep Fast Vision offers a mix of automatic, semi-automatic, and user-defined parameters. The following table 1 provides a summary of these parameters:

\begin{table}[h]
\centering
\caption{Parameters for Deep Fast Vision}
\begin{tabularx}{\textwidth}{lXl}
\toprule
\textbf{Category} & \textbf{Parameters} & \textbf{Automation Level} \\
\midrule
Automatic & Loss function, Identify Train-Val and Test folders, Establish output layer size and activation functions, Calculate and apply class weights, Generate appropriate data generators for train, val, and test data, Resize images to transfer model specification, Retrieve transfer architecture's preprocessing function, Prepare data augmentation, Pre-train dense layers, Monitor and load optimal weights, Conduct test, Create validation curves, Produce confusion matrices, Provide model architecture summary & Automatic \\
\midrule
Semi-automatic & Dropout between dense layers, Augmentation settings for training data, Unfreeze and train transfer model, Load best weights based on validation results, Test after loading the best weights & Adjustable \\
\midrule
User-defined & Data paths, Transfer learning architecture, Dense layer configuration, Callbacks, Transfer model unfreeze blocks & User-defined \\
\bottomrule
\end{tabularx}
\end{table}

\subsubsection*{Installation \& Documentation}

To install the library, you can use the pip package manager. Open a terminal and type the following:

\begin{lstlisting}[language=bash]
pip install deepfastvision
\end{lstlisting}

For those intending to leverage the older 'tensforflow-gpu', the following variant should be used:

\begin{lstlisting}[language=bash]
pip install deepfastvision[gpu]
\end{lstlisting}

The official documentation for the library provides comprehensive guides and notes to get you started. It covers everything from basic usage to advanced functionalities.

You can access the documentation at:
\begin{verbatim}
https://fabprezja.github.io/deep-fast-vision/
\end{verbatim}

For specific modules and functions, the API reference section can be particularly helpful. For hands-on practice, the GitHub homepage offers a variety of tutorials and examples. 

Official GitHub:
\begin{verbatim}
https://github.com/fabprezja/deep-fast-vision
\end{verbatim}

Whether you are a beginner or an advanced user, the documentation is designed to provide you with all the information you need to make the most of the library. If you encounter any issues or bugs, you can report them on the library's GitHub repository.

\section*{Code Examples}
In our exploration of the deepfastvision library, we identified two distinct levels of abstraction: low and high. These levels dictate how users interact with the library and the degree of control they have over specific processes. Bellow we find results related to those levels of abstraction.

\subsubsection*{High Level of Abstraction}
\begin{lstlisting}[language=Python]
import wandb
from wandb.keras import WandbCallback
from deepfastvision.core import DeepTransferClassification

# Initialize wandb
wandb.init(project='your_project_name', entity='your_username')

# Create a DeepTransferClassification object
experiment = DeepTransferClassification(paths={'train_val_data': 'path_to_train_val_data',
                                               'test_data_folder': 'path_to_test_data'},
                                        saving={'save_weights_folder':'path_to_save_weights'},
                                        model= {'transfer_arch': 'VGG16',
                                                'dense_layers': [144,89,55],
                                                'unfreeze_block': ['cblock5']},
                                        training={'epochs': 15,
                                                  'learning_rate': 0.0001,
                                                  'metrics': ['accuracy'],
                                                  'callback': [WanDBCallback()]})

model, results = experiment.run()
\end{lstlisting}

Upon execution, the code returns the trained model, a results dictionary detailing training and evaluation outcomes along with model configurations, validation curves, and a confusion matrix, both tailored to the target type and given labels and metrics. The user supplies data paths from which the library sources and loads training, validation, and testing data. Additionally, the library incorporates the VGG16\cite{simonyan2014very} pre-trained model for transfer learning, integrates dense layers with neuron counts of 144, 89, and 55, and integrates any user-specified callback, such as WanDB\cite{wandb}.

\subsubsection*{Low Level of Abstraction}
\begin{lstlisting}[language=Python]
from deepfastvision.core import DeepTransferClassification

experiment = DeepTransferClassification(paths={
    'train_val_data': 'path_to_train_val_data',
    'test_data_folder': 'path_to_test_data',
    'external_test_data_folder': 'path_to_external_test_data',
},
model={
    'image_size': (224, 224),
    'transfer_arch': 'VGG19',
    'pre_trained': 'imagenet',
    'before_dense': 'Flatten',
    'dense_layers': [610, 377, 233, 144, 89, 55],
    'dense_activations': 'elu',
    'initializer': 'he_normal',
    'batch_norm': True,
    'regularization': 'Dropout+L2',
    'l2_strength': 0.001,
    'dropout_rate': 0.35,
    'unfreeze_block': ['cblock1', 'cblock2', 'cblock5'],
    'freeze_up_to': 'flatten',
},
training={
    'epochs': 9,
    'batch_size': 32,
    'learning_rate': 2e-5,
    'optimizer_name': 'Adam',
    'add_optimizer_params': {'clipnorm': 0.8},
    'class_weights': True,
    'metrics': ['accuracy', 'recall', 'precision'],
    'augmentation': 'custom',
    'custom_augmentation':[user_function]
    'callback': [WandbCallback(), learning_rate_schedule],
    'early_stop': 0.20,
    'warm_pretrain_dense': True,
    'warm_pretrain_epochs': 9,
},
evaluation={
    'auto_mode': True,
},
saving={
    'save_weights': True,
    'save_weights_folder': 'path_to_save_weights',
    'save_best_weights': 'val_loss',
},
misc={
    'show_summary': True,
    'plot_curves': True,
    'show_min_max_plot': True,
    'plot_conf': True,
})

model, results = experiment.run()
\end{lstlisting}

In this detailed low-level of abstraction configuration, users are granted enhanced control over the model's architecture and training process. This configuration encompasses a blend of user-specified modifications and the default settings. Contrasting with the medium abstraction level, the user has introduced a \texttt{'external\_test\_data\_folder'} for supplementary test data, enabled batch normalization, added an extensive set of dense layers (610, 377, 233, 144, 89, 55), and set regularization to \texttt{'Dropout+L2'} with an L2 strength of 0.001. Further, the user has decided which blocks to unfreeze, namely \texttt{'block1'}, \texttt{'block2'}, and \texttt{'block5'}, while freezing layers up to the \texttt{'flatten'} phase. The augmentation is customized by the user, and additional callbacks, WandbCallback and \texttt{learning\_rate\_schedule}, have been incorporated. There's also an activation of early stopping with a threshold set at 0.20 relative to the total epochs, an increase in epochs to 25, and a warm pretraining for dense layers spanning 9 epochs. 

By default, the weight initializer is \texttt{'he\_normal'}\cite{he2015delving}, class weights are active, and the image size is defined at (224, 224). The pre-trained weights are sourced from \texttt{'imagenet'}\cite{deng2009imagenet}, the preceding layer before dense layers is \texttt{'Flatten'}, and the dense layer activations utilize \texttt{'elu'}\cite{clevert2015fast}. The learning rate stands at 2e-5, with the evaluation configuration's \texttt{'auto\_mode'} set to true, implying automatic evaluation of the best weights. The model saves its most optimal weights based on \texttt{'val\_loss'}, and several other features, such as displaying the model summary, plotting various curves, showing a min-max plot, and representing data through a confusion matrix, are also activated. As observed in prior examples, all automation processes remain in effect.

\subsection*{Model Prediction}
\begin{lstlisting}[language=Python]
predictions = experiment.model_predict('folder_path')
\end{lstlisting}
The \texttt{model\_predict} method uses the trained model to predict all images in a given folder. The method returns image, path, predicted label, confidence, and variance for each image in the folder. It can be sorted by variance (across labels) for identifying confusing instances or by metric (e.g., accuracy).

\subsection*{Export Results and Model}
\begin{lstlisting}[language=Python]
experiment.export_all(results, base_path='folder_path_to_results', export_model=True, additive=True)
\end{lstlisting}
The \texttt{export\_all} method exports all results, best weights, and the trained model into a folder. With \texttt{additive=True}, the user may iterate the experiment and obtain results in new randomly named folders.

\subsection*{Extract Features}
\begin{lstlisting}[language=Python]
X_train, y_train, X_val, y_val, X_test, y_test, X_test_external, y_test_external = experiment.model_feature_extract(layer_index=None, layer_name='my_layer')
\end{lstlisting}
The \texttt{model\_feature\_extract} method can be used to extract features from any layer in the model while respecting the used train, val, test(s) indices.

\subsection*{Run Example}
\begin{figure}[!ht]
\centering
\includegraphics[scale=0.36]{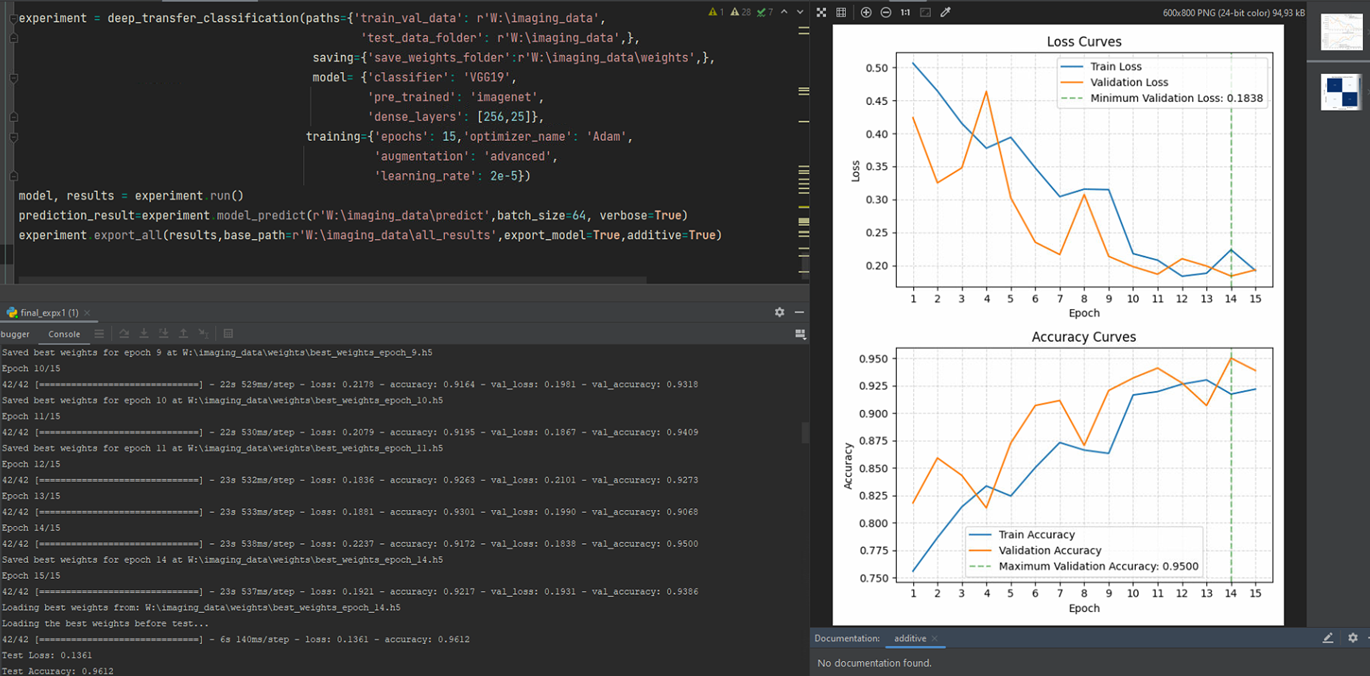}
\caption{An example of executing Deep Fast Vision on an image dataset.}
\label{runexp}
\end{figure}

In Figure \ref{runexp}, a standard example of a run is depicted, showcasing the plots and test results that are generated automatically.

\section*{Discussion}
The advent of deep learning has undeniably transformed the landscape of various industries, with vision-based applications standing at the forefront of this transformation. As the Introduction section highlighted, while the potential of deep learning in visual tasks is immense, several challenges persist. From the complexity of current frameworks to the scarcity of extensive labeled datasets, these challenges can pose significant roadblocks for many researchers and developers.Transfer learning, with its promise to leverage knowledge from larger datasets to improve performance on smaller datasets, has been viewed as a potential solution to some of these challenges.

The "Deep Fast Vision" library is currently utilized in research studies\cite{pastfut,prezja2023improving,aug,avt}. At its core, the library seeks to simplify the intricacies of deep learning frameworks, making it more accessible to a broader audience. Its strength lies in its ability to streamline the prototyping process, as seen in the Methods section, where both high and low levels of abstraction were explored. By offering a blend of automatic, semi-automatic, and user-defined parameters, the library provides users with the flexibility to tailor their approach based on their expertise and requirements. Comparing "Deep Fast Vision" with AutoKeras, another prominent Auto-ML solution, highlights some of the unique offerings of the former. While AutoKeras excels in identifying suitable deep learning models and optimizing hyperparameters, "Deep Fast Vision" distinguishes itself with its focus on transfer learning for vision tasks. Its foundation on TensorFlow and Keras ensures that users benefit from the robustness of these leading frameworks while enjoying a more streamlined experience tailored for rapid prototyping.

The installation and documentation details underscore the library's commitment to user-friendliness. From easy installation commands to comprehensive documentation and a dedicated GitHub repository, "Deep Fast Vision" ensures that users have all the resources they need to navigate the library effectively. The Code Examples section further exemplifies the versatility of the library. Whether a user prefers a high level of abstraction, where many of the processes are automated, or a low level of abstraction, which offers more granular control, "Deep Fast Vision" caters to both preferences.

However, like all tools, "Deep Fast Vision" is not without its limitations. While its focus on transfer learning for vision tasks is key, there might be scenarios where specialized solutions are required. Additionally, the library's dependency on TensorFlow and Keras could pose challenges for those who prefer other deep learning frameworks.

\bibliography{main}

\section*{Acknowledgements}
The authors extend their sincere gratitude to Leevi Annala, Rodion Enkel, Sampsa Kiiskinen, Suvi Lahtinen, Leevi Lind, and Kimmo Riihiaho.

\section*{Data Availability}
All code materials are available at the official GitHub repository.

\section*{Author contributions statement}
Conceptualization: F. P.;
Methodology: F. P.; 
Data Curation: F. P.;
Writing – review \& editing: F. P.;

\section*{Additional information}
 \textbf{Competing interests}
 All authors declare that they have no conflicts of interest.

\end{document}